%% file: main.tex
\documentclass[9pt,twoside]{article}
\input{manuscript-config.tex}
\input{revision-style.tex}

\begin{document}
\MakePerspectiveTitle
\begin{multicols}{2}
\input{content.tex}
\end{multicols}
\end{document}

%% file: manuscript-config.tex
\newif\ifanonymous
\newif\ifarxiv
\anonymousfalse
\arxivtrue

\usepackage[paperwidth=216mm,paperheight=285mm,top=14mm,bottom=16mm,left=15mm,right=15mm,headheight=18.5pt,headsep=3mm,footskip=8mm]{geometry}
\usepackage[T1]{fontenc}
\usepackage[utf8]{inputenc}
\usepackage{mathpazo}
\usepackage[scaled=0.92]{helvet}
\usepackage{microtype,xcolor,graphicx,needspace,booktabs,tabularx,amsmath,amssymb,multicol,titlesec,fancyhdr,float,etoolbox,environ}
\usepackage[hang,flushmargin]{footmisc}
\usepackage{marvosym}
\usepackage[font=small,labelfont=bf,labelsep=period]{caption}

\usepackage[authoryear,round,sort]{natbib}
\usepackage[colorlinks=true,linkcolor=black,citecolor=black,urlcolor=blue!55!black]{hyperref}
\usepackage[nameinlink]{cleveref}
\crefname{figure}{Fig.}{Figs.}
\Crefname{figure}{Fig.}{Figs.}

\AtBeginEnvironment{thebibliography}{\fontsize{7.4}{8.8}\selectfont\setlength{\itemsep}{0pt}\setlength{\parsep}{0pt}}
\makeatletter
\providecommand{\abx@aux@refcontext}[1]{}
\providecommand{\abx@aux@cite}[2]{}
\providecommand{\abx@aux@segm}[3]{}
\providecommand{\abx@aux@page}[2]{}
\providecommand{\abx@aux@read@bbl@mdfivesum}[1]{}
\providecommand{\abx@aux@defaultrefcontext}[3]{}
\makeatother

\definecolor{Accent}{HTML}{497FC4}
\titleformat{\section}{\normalfont\bfseries\fontsize{11}{13}\selectfont}{\thesection}{0.55em}{}
\titlespacing*{\section}{0pt}{11pt plus 2pt minus 1pt}{4pt}
\titleformat{\subsection}{\normalfont\bfseries\fontsize{9.5}{11.5}\selectfont}{\thesubsection}{0.55em}{}
\titlespacing*{\subsection}{0pt}{8pt plus 2pt minus 1pt}{3pt}

\newcommand{\ArticleType}{Perspective}
\newcommand{\JournalName}{J Intel Connect Veh}
\newcommand{\PaperTitle}{Rethinking World Models for Safety-Critical Embodied Systems}

\newcommand{\AuthorLine}{Kailang Ma\textsuperscript{1}, Heye Huang\textsuperscript{1,\Letter}, Inhi Kim\textsuperscript{1}, Kitae Jang\textsuperscript{1}}
\newcommand{\AffiliationLine}{\textsuperscript{1}Cho Chun Shik Graduate School of Mobility, Korea Advanced Institute of Science and Technology, Daejeon 34051, South Korea.}
\newcommand{\CorrespondingAuthor}{\textsuperscript{\Letter}\ Corresponding author. E-mail: \href{mailto:heye.huang@kaist.ac.kr}{heye.huang@kaist.ac.kr}}
\newcommand{\ManuscriptHistory}{}

\ifanonymous
  \ifarxiv
    \hypersetup{pdftitle={\PaperTitle},pdfauthor={},pdfsubject={Preprint},pdfkeywords={}}
  \else
    \hypersetup{pdftitle={\PaperTitle},pdfauthor={},pdfsubject={Perspective},pdfkeywords={}}
  \fi
\else
  \ifarxiv
    \hypersetup{pdftitle={\PaperTitle},pdfauthor={Kailang Ma, Heye Huang, Inhi Kim, Kitae Jang},pdfsubject={Preprint},pdfkeywords={}}
  \else
    \hypersetup{pdftitle={\PaperTitle},pdfauthor={Kailang Ma, Heye Huang, Inhi Kim, Kitae Jang},pdfsubject={Perspective, Journal of Intelligent and Connected Vehicles},pdfkeywords={}}
  \fi
\fi

\ifarxiv
  \renewcommand{\headrulewidth}{0pt}
  
\else
  \renewcommand{\headrulewidth}{0.35pt}
  \renewcommand{\headrule}{\hbox to\headwidth{\rule{0.965\headwidth}{\headrulewidth}\hfill}}
  
\fi

\fancypagestyle{firstpage}{%
  \fancyhf{}%
  \fancyhead[R]{\sffamily\fontsize{8}{9}\selectfont\thepage}%
  \fancyfoot{}%
  \ifarxiv
    \renewcommand{\headrulewidth}{0pt}%
  \else
    \renewcommand{\headrulewidth}{0.35pt}%
    \renewcommand{\headrule}{\hbox to\headwidth{\rule{0.965\headwidth}{\headrulewidth}\hfill}}%
  \fi
}

\newcommand{\MakePerspectiveTitle}{%
  \thispagestyle{firstpage}%
  \noindent\begin{minipage}{\textwidth}
  \ifarxiv
  \else
    {\sffamily\bfseries\itshape\color{Accent}\fontsize{14}{16}\selectfont\JournalName\par}
    \vspace{3pt}
    \noindent\hfill{\sffamily\color{Accent}\fontsize{9}{11}\selectfont\ArticleType}\par\vspace{3pt}
  \fi
  {\raggedright\sffamily\bfseries\fontsize{18}{21}\selectfont\PaperTitle\par}
  \ifanonymous
    \vspace{3pt}
  \else
    \vspace{4pt}
    {\raggedright\fontsize{9.5}{12}\selectfont\AuthorLine\par}
    \ifarxiv
      \vspace{5pt}
      {\raggedright\sffamily\fontsize{7.5}{9.5}\selectfont\AffiliationLine\par}
      \vspace{2pt}
      {\raggedright\sffamily\fontsize{7.5}{9.5}\selectfont\CorrespondingAuthor\par}
    \fi
  \fi
  \end{minipage}%
  \ifarxiv\vspace{6mm}\else\vspace{8mm}\fi
  \ifanonymous
  \else
    \ifarxiv
    \else
      \begingroup\renewcommand{\thefootnote}{}
      \footnotetext{\noindent\rule{0.965\textwidth}{0.35pt}\par\vspace{3pt}
      \noindent\raggedright\sffamily\fontsize{6.7}{8.2}\selectfont
      \AffiliationLine\\
      \CorrespondingAuthor
      \ifx\ManuscriptHistory\empty\else\\\ManuscriptHistory\fi}
      \endgroup
    \fi
  \fi}

\makeatletter
\newif\ifRIWM@discarding
\RIWM@discardingfalse
\let\RIWM@original@section\section
\let\RIWM@original@input\input

\ifanonymous
  \renewcommand{\section}{\@ifstar{\RIWM@section@star}{\RIWM@original@section}}
  \newcommand{\RIWM@section@star}[1]{%
    \ifRIWM@discarding
      \egroup
      \global\RIWM@discardingfalse
    \fi
    \ifstrequal{#1}{Author contributions}{%
      \global\RIWM@discardingtrue
      \setbox\z@=\vbox\bgroup
    }{%
      \ifstrequal{#1}{Acknowledgements}{%
        \global\RIWM@discardingtrue
        \setbox\z@=\vbox\bgroup
      }{%
        \RIWM@original@section*{#1}%
      }%
    }%
  }
  \pretocmd{\bibliography}{%
    \ifRIWM@discarding
      \egroup
      \global\RIWM@discardingfalse
    \fi
  }{}{}
  \renewcommand{\input}[1]{%
    \ifRIWM@discarding
      \egroup
      \global\RIWM@discardingfalse
    \fi
    \RIWM@original@input{#1}%
  }
\fi
\makeatother

\ifanonymous
  \NewEnviron{biography}[1]{}
\else
  \newenvironment{biography}[1]{\par\smallskip\noindent\begin{minipage}{\columnwidth}\begin{minipage}[t]{0.20\columnwidth}\vspace{0pt}\includegraphics[width=\linewidth]{#1}\end{minipage}\hfill\begin{minipage}[t]{0.77\columnwidth}\vspace{0pt}\fontsize{8}{9.5}\selectfont}{\end{minipage}\end{minipage}\par\smallskip}
\fi

%% file: revision-style.tex
\usepackage[normalem]{ulem}
\definecolor{RevisionBlue}{RGB}{0,65,155}
\definecolor{RevisionRed}{RGB}{165,30,35}
\ifarxiv
  \newcommand{\added}[1]{#1}
  \newcommand{\deleted}[1]{}
  \newenvironment{revisiontext}{\par\begingroup}{\par\endgroup}
\else
  \newcommand{\added}[1]{{\color{RevisionBlue}#1}}
  \newcommand{\deleted}[1]{{\color{RevisionRed}\sout{#1}}}
  \newenvironment{revisiontext}{\par\begingroup\color{RevisionBlue}}{\par\endgroup}
\fi

%% file: content.tex
\section{Introduction}

World-model research has progressed from compact latent dynamics for control \citep{ha2018world} to generative, controllable, and interactive simulators of embodied environments. GAIA-2 \citep{russell2025gaia2} advances controllable multi-view synthesis; UniSim \citep{yang2024unisim} enables interactive real-world simulation; and TrafficBots \citep{zhang2023trafficbots}, OccWorld \citep{zheng2024occworld}, and DriveDreamer4D \citep{zhao2025drivedreamer4d} extend multi-agent behavior modeling, 3D occupancy prediction, and 4D scene representation. These advances extend generative and predictive capabilities, but do not necessarily improve a model's understanding of what matters for safety: even an accurate and visually realistic model may overlook evidence that should prompt a change in the current action.

\added{\label{rev:scope}Driving combines partial observability, interaction, and consequential real-time decisions, making it our running example; a humanoid example illustrates broader application.} Consider an unprotected left turn. An unobstructed crossing may be the most likely future, whereas a pedestrian emerging from behind an occluder, despite its lower probability, may determine whether acceleration is defensible. Waiting, creeping, and accelerating also alter how other road users respond, thereby changing both available information and recovery space. For safety-critical embodied systems, the key question is therefore not only what is likely to happen, but what could change the action, how alternative actions alter consequences, and whether accumulated experience should revise the current judgment. Objectives dominated by likelihood, reconstruction, or visual fidelity may suppress precisely this decision-relevant evidence.

\begin{figure}[H]
\centering
\includegraphics[width=\textwidth,keepaspectratio]{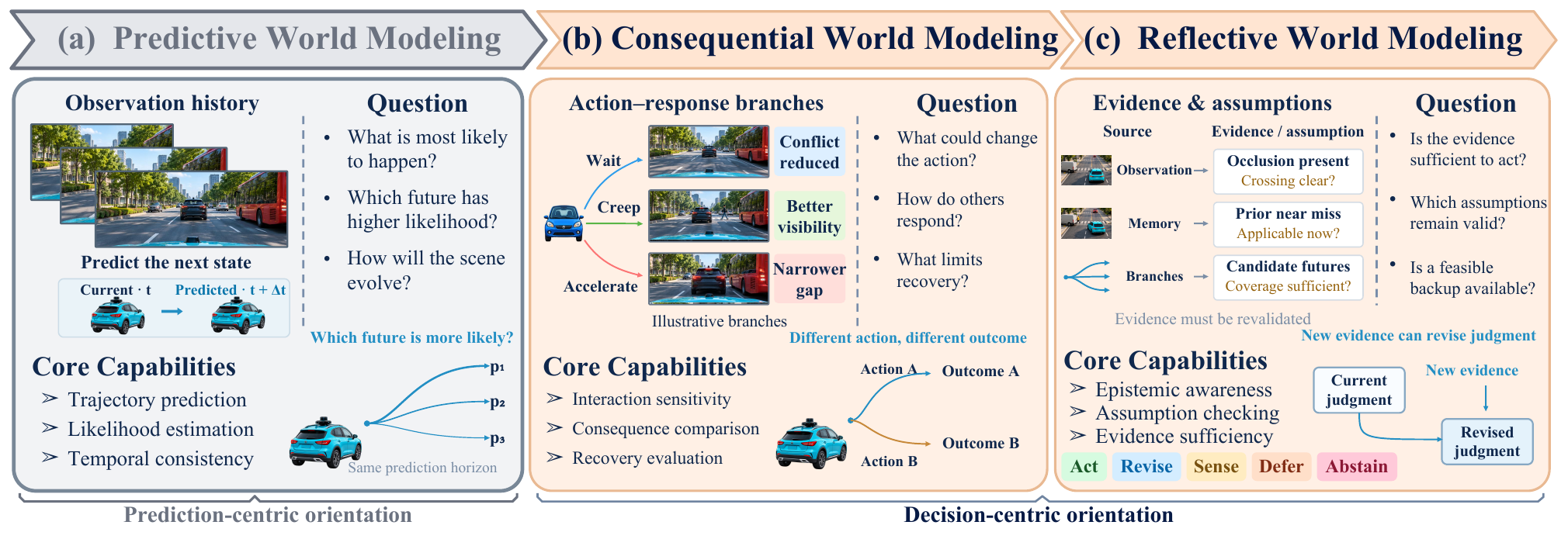}
\caption{Cognitive progression across predictive, consequential, and reflective world modeling, from a prediction-centric to a decision-centric orientation.}
\label{fig:cognitive-progression}
\end{figure}

We therefore organize world-model capability as a three-level cognitive progression. As shown in \cref{fig:cognitive-progression}, predictive modeling asks what is most likely to happen; consequential modeling asks what could change the action; and reflective modeling asks whether the available evidence is sufficient to support action. These capabilities are cumulative rather than mutually exclusive. We define consequential futures as credible branches that alter the feasibility, acceptability, or recoverability of \mbox{candidate actions}.

Building on this view, we propose the Risk-Informed World Model (RIWM) as a research direction for safety-critical embodied systems. RIWM organizes world modeling around action consequences: decision-relevant evidence shapes representation, counterfactual reasoning, episodic memory, and runtime safety assurance, while execution feedback continuously revises that evidence. It treats physical, social, and operational consequences as distinct decision domains, with epistemic uncertainty qualifying the evidence used to assess them. Risk-informed modeling does not imply eliminating risk, achieving complete causal identification, or providing unconditional safety guarantees. Instead, consequences, epistemic uncertainty, and recoverability should materially influence what the model represents, reasons about, remembers, and when it should act, revise, sense, defer, or abstain. This direction extends efforts toward unified environmental understanding, decision making, and control in autonomous driving \citep{wang2021level5}, while targeting safety-critical embodied systems more broadly.

\section{Three structural mismatches}

\subsection{Likelihood-risk mismatch}

The futures that matter most for action are not necessarily those that occur most often. As scenario dimensionality and interaction complexity increase, safety-critical events occupy a rapidly shrinking fraction of available data, \added{a pattern known as} the curse of rarity in autonomous-driving safety research \citep{liu2024curse}. Safety validation must therefore prioritize informative and discriminative scenarios rather than rely on inefficient random exposure \citep{feng2023dense}. As the unprotected-left-turn example shows, a lower-probability but consequential credible branch may govern the decision even when it is not the most likely future. Probability alone therefore cannot express a branch's importance to the current action. A scalar probability-severity score similarly obscures important decision structure. Equivalent scores may correspond to different interaction responses, threatened requirements, or recovery margins, yet these distinctions determine whether the system should act, revise, sense, defer, or abstain.

This limitation becomes more consequential under epistemic uncertainty. Aleatoric uncertainty concerns irreducible variability or observation noise \citep{kendall2017uncertainties}, whereas epistemic uncertainty reflects limited knowledge about the model or data-generating process and may be reducible with additional evidence \citep{hullermeier2021uncertainty}. Crucially, epistemic uncertainty may mean that the relevant branch has not been represented at all. Once occlusion topology, conflicting actors, or recovery space has been omitted or collapsed in the learned state, downstream risk scoring cannot reliably reconstruct it. Safety relevance must therefore shape representation and branch allocation rather than be appended after prediction.

\end{multicols}\begin{multicols}{2}

\subsection{Prediction-intervention mismatch}

An embodied agent does not merely predict the future; its actions help create it. At the left-turn intersection, waiting, creeping, and accelerating alter not only the ego trajectory but also others' responses, available information, and recovery space. The future is therefore not a policy-independent sequence, but a branching structure jointly shaped by action, interaction, information change, \mbox{and consequence.}

Several levels of reasoning should remain distinct. Action-conditioned prediction estimates what may follow a candidate action; intervention-sensitive rollout models how responses vary with that action; counterfactual comparison contrasts the resulting action-response-consequence chains; and causal identification requires explicit causal assumptions \citep{pearl2009causality}. Interactive prediction and multi-agent simulation \deleted{model how futures vary with candidate actions} \added{model dependencies among agents' future trajectories} \citep{huang2023gameformer,sun2022m2i,suo2021trafficsim}. \deleted{but }\added{However,} action conditioning within a predictive or simulated model does not by itself identify how the world would respond to an unchosen intervention. RIWM therefore requires reliable action-conditioned rollout and bounded counterfactual comparison, while making the assumptions behind stronger causal \mbox{claims explicit.}

\subsection{Horizon-accumulation mismatch}

A sequence of locally acceptable decisions can still accumulate into an unsafe outcome. A mildly assertive merge may remain collision-free while triggering a downstream braking cascade, whereas repeated conservative waiting may avoid immediate conflict yet produce deadlock or provoke less predictable behavior. Frame-wise safety therefore does not guarantee \mbox{long-horizon recoverability.}

This mismatch cannot be resolved simply by extending the prediction horizon or context window. Longer prediction looks farther ahead within a rollout, whereas evidence continuity preserves why an event mattered across successive observations, actions, and feedback. \added{The system should therefore retain the consequential structure of experience rather than indiscriminate historical detail. This includes what action was taken, how the environment responded, what consequence followed, which assumptions failed, and whether recovery succeeded.} Existing agent-memory and lifelong-learning research provides mechanisms for retaining and revising experience over time \citep{meng2025lifelong,park2023generative}, but such mechanisms do not by themselves preserve its safety relevance. RIWM therefore requires failures, near misses, and recoveries to be retained with their consequences, provenance, failed assumptions, and recovery outcomes so that they can revise current representation and future branching.

\section{The risk-informed world model}

\begin{figure}[H]
\centering
\includegraphics[width=\textwidth,keepaspectratio]{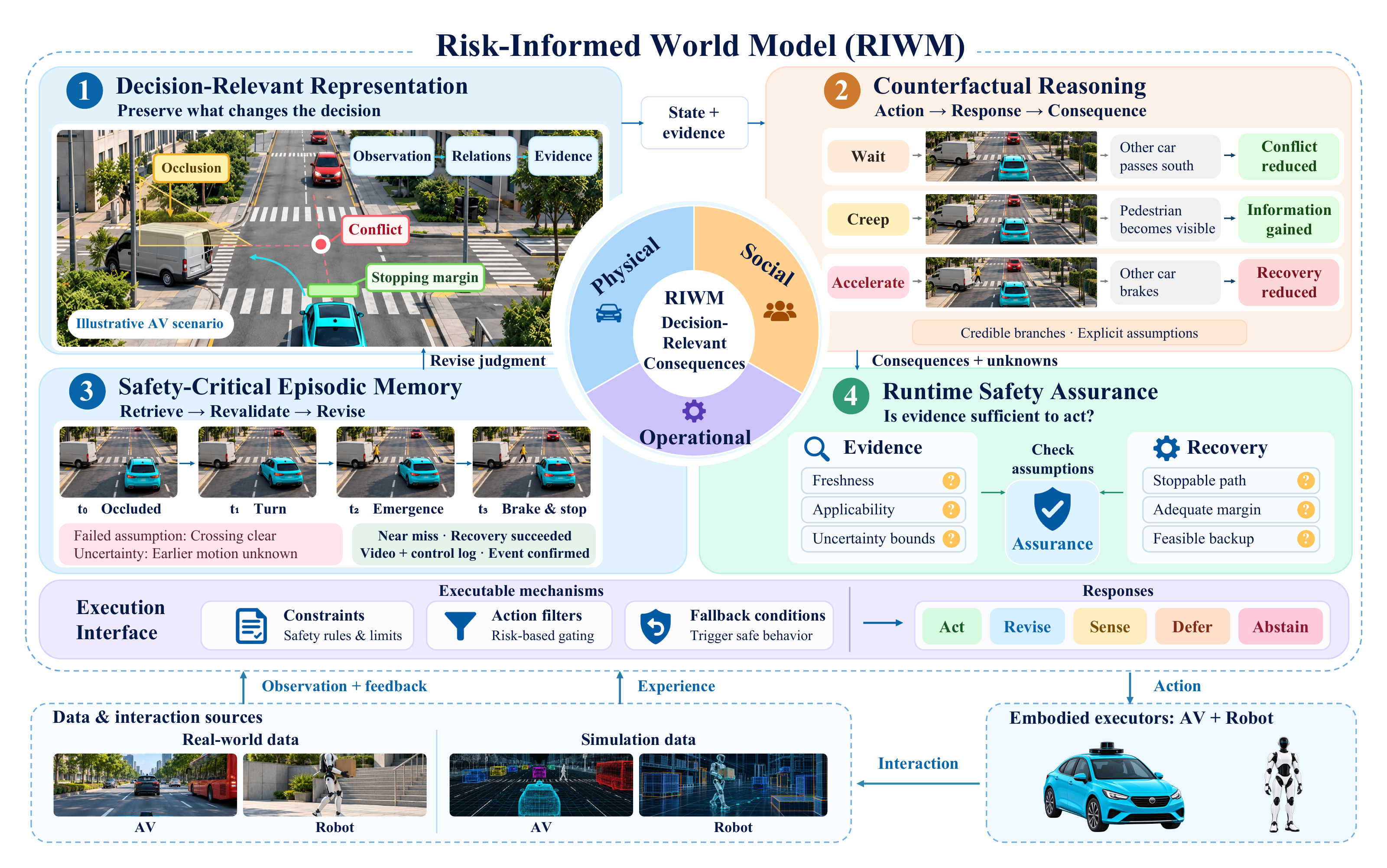}
\caption{Conceptual architecture of a risk-informed world model for safety-critical embodied systems.}
\label{fig:riwm-architecture}
\end{figure}

The three structural mismatches require world models to organize evidence around consequences, intervention, uncertainty, and recovery. RIWM organizes consequences into physical (collision and recovery space), social (norms, interactions, and responses), and operational (progress, efficiency, information gain, and deadlock) domains. RIWM comprises four interdependent capabilities: decision-relevant representation preserves information that may change the action boundary; counterfactual reasoning compares responses and consequences under candidate actions; safety-critical episodic memory uses past failures, near misses, and recoveries to revise current judgment; and runtime safety assurance evaluates whether available evidence is sufficient to support action under explicit assumptions. \Cref{fig:riwm-architecture} illustrates the closed-loop interaction between the model and the environment, integrating these capabilities with decision-relevant consequences, epistemic uncertainty, and execution feedback. Temporally, episodic memory looks backward, counterfactual reasoning unfolds forward, and runtime safety assurance operates concurrently with action, while epistemic uncertainty constrains the strength of evidence used to judge consequences.

\end{multicols}\begin{multicols}{2}

\begin{revisiontext}\label{rev:boundary}
A world model qualifies as an RIWM if it (i) predicts action-conditioned consequences and recovery, (ii) is learned and updated in a decision-sensitive way, and (iii) exposes uncertainty and unsupported premises to downstream assurance. These requirements distinguish RIWM from simply combining existing prediction, reasoning, memory, and assurance methods. RIWM draws on these methods but organizes their interaction around what could change a decision: predicted consequences guide action comparison, relevant memories revise the model's hypotheses, and uncertainty and recovery estimates inform runtime assurance. Causal reasoning clarifies the assumptions under which action-conditioned comparisons support intervention claims. In this sense, RIWM complements risk-aware planning \citep{cheng2026dual} by shaping the world model on which planning relies.
\end{revisiontext}

\subsection{Decision-relevant representation}

Decision relevance concerns whether information changes the feasibility, acceptability, recoverability, ranking, or evidential support of candidate actions, rather than its visual salience or frequency. Decision-relevant representation consequently prioritizes relations that support action comparison. In driving, these include conflict regions, occlusion, right of way, social norms, escape space, uncertainty in others' responses, and the model's \mbox{knowledge boundary.}

\begin{revisiontext}\label{rev:definition}
This relevance can be tested by varying evidence while holding the task, candidate actions, horizon, and requirements fixed. A credible branch is consequential if including it changes action ranking, admissibility, or the evidence required for execution. Branch-removal tests must specify how its probability mass is reassigned.
\end{revisiontext}

Existing work provides partial but complementary ingredients for decision-relevant representation. K-Risk \citep{huang2026krisk} links structured driving trajectories and event-level metadata with extracted high-risk events and \added{large language model (LLM)}-generated semantic annotations, while RiskNet \citep{liu2026risknet} represents directional interaction risk. However, neither component alone determines which evidence changes an action boundary across physical, social, and operational consequences. These consequence domains may coexist and need not be compressed into a single cost; instead, their distinct effects on action selection should remain identifiable. Epistemic uncertainty is not a consequence domain but qualifies the evidence supporting consequence assessment and action selection.

Two futures may have similar pixels and short-term trajectories yet differ decisively in recoverability when a guardrail and adjacent vehicle remove lateral escape space. Representation should thus be evaluated not only by image, occupancy, or trajectory accuracy, but also by whether changes in occlusion, right of way, escape space, and uncertainty revise counterfactual branches and action ranking. For safety-critical embodied systems, the representational objective should therefore shift from preserving everything accurately to preserving what may change the decision. \added{This can be implemented by complementing predictive learning with decision-related objectives.} Consequence, recovery margin, and evidence sufficiency should shape representation before execution, rather than being appended as a downstream risk score.

\begin{revisiontext}\label{rev:learning}
One possible implementation uses a shared history encoder and an action-conditioned latent model with risk, value, or recovery prediction heads. An illustrative joint objective is
\[
\mathcal L =
\underbrace{\mathcal L_{\mathrm{dyn}}}_{\substack{\text{Dynamics}\\\text{prediction}}}
+\underbrace{\lambda_v\mathcal L_{\mathrm{value}}}_{\substack{\text{Value}\\\text{estimation}}}
+\underbrace{\lambda_r\mathcal L_{\mathrm{risk}}}_{\substack{\text{Risk}\\\text{estimation}}}
+\underbrace{\lambda_m\mathcal L_{\mathrm{rec}}}_{\substack{\text{Recovery}\\\text{margin}}}.
\]
Each term measures the corresponding prediction or estimation error; the weights balance their contributions to the shared representation. Depending on available supervision, training can use expert-action imitation, auxiliary risk targets computed from traffic-risk measures \citep{wang2025riskprediction}, or reinforcement learning through reward and value objectives, as in TD-MPC2 \citep{hansen2024tdmpc2}. Simulation or intervention records could supply constraint-cost and recovery targets. Recovery should be assessed relative to a specified backup controller; oversampling rare events requires probability recalibration.
\end{revisiontext}

\subsection{Counterfactual reasoning}

Exhaustive simulation of all futures is neither feasible nor necessary. Rollout breadth, horizon, and fidelity should instead adapt to decision relevance. Familiar, weakly interactive situations with ample recovery margin may require only compact prediction, whereas severe occlusion, asymmetric responses, relevant failure memories, or narrowing recovery space should trigger broader, longer, or more detailed rollouts. Modeling resources should therefore be allocated dynamically rather than uniformly.

Under a fixed branch budget, counterfactual reasoning should preserve both the most likely baseline and credible alternatives that may change the safety boundary. Such alternatives should be supported by current observation, calibrated uncertainty, valid episodic memory, or bounded perturbation rather than arbitrary hazard generation. Generability is not credibility. Branch coverage should therefore make explicit which futures were considered, what evidence supports them, and which critical variables remain unknown.

Existing approaches provide partial foundations for such comparisons. Human-aware planning represents how ego actions may influence human responses \citep{sadigh2016planning}, while LEAD \citep{huang2025lead} adapts game-theoretic decisions to changing interactions. These formulations support action-dependent comparison but do not by themselves establish that the resulting branches are causally identified or sufficiently complete for safety-critical decisions. RIWM should therefore concentrate finite computation on credible, decision-relevant alternatives and \added{support acting, revising, sensing, deferring, or abstaining when evidence is insufficient.}

\begin{revisiontext}\label{rev:allocation}
Relevance is progressively estimated as planning and execution unfold. Within a planning cycle, coarse geometry and short rollouts guide selective branch expansion. RISE \citep{lu2026rise} estimates the planning gain from further rollout to adapt its depth. Across execution steps, new observations support pruning, new hypotheses, and replanning; AdaReP \citep{cheng2026adarep} adapts replanning to prediction mismatch and local dynamics sensitivity. Reserving computation for exploration helps revisit overlooked branches. Before an irreversible action, unresolved critical uncertainty should prompt further sensing or a feasible waiting or backup strategy.
\end{revisiontext}

\begin{revisiontext}\label{rev:vlm}
Vision-language models (VLMs) and LLMs can provide System~2-style semantic hypotheses for world-model assessment. AutoVLA \citep{zhou2025autovla} integrates semantic reasoning and trajectory generation in a single autoregressive model. A possible collaboration within RIWM is for VLMs/LLMs to propose hypotheses, vision-language-action models (VLAs) or planners to propose actions, and world-model rollouts to assess consequences. For an occluded-pedestrian hypothesis, the world model could use visibility and motion information to assess braking and stopping margin. The hypothesis's probability still requires observational or calibrated predictive evidence. Deliberation can be invoked within a latency budget while a fast controller maintains feasible control.
\end{revisiontext}

\subsection{Safety-critical episodic memory}

Agent cognitive architectures distinguish memory stores and operations \citep{sumers2024cognitive}, and generative agents demonstrate retrieval and reflection over recorded experience \citep{park2023generative}. Lifelong robotic learning additionally studies how knowledge can be preserved and combined over time \citep{meng2025lifelong}. RIWM builds on these foundations but requires more than a history indexed by visual similarity: a reusable safety episode should link the situation, action, others' responses, consequence, recovery, assumptions, confidence, provenance, and time. Retention should prioritize failures, near misses, successful recoveries, and other experiences with high consequence, surprise, recovery, or transfer value.

Retrieval should emphasize interaction structure and action-response relations rather than appearance alone. Equally important, memory must remain revisable: changes in sensors, policies, traffic rules, or operating conditions may invalidate earlier experience. Memories should therefore support conflict tracking, revalidation, down-weighting, or forgetting under distribution shift. In RIWM, consequential experience is not merely training data but auditable evidence that can revise current interpretation, counterfactual branching, and recovery strategy at runtime.

\subsection{Runtime safety assurance}

World-model outputs should inform runtime safety assurance but remain distinct from constraint enforcement. \added{Model predictive control (MPC)} \citep{rawlings2020mpc}, \added{control barrier functions (CBFs)} \citep{ames2019cbf}, reachability analysis \citep{althoff2021set}, and runtime assurance \citep{hobbs2023runtime} provide \deleted{executable mechanisms for enforcing safety constraints} \added{complementary tools for constraint handling, safety verification, and runtime filtering}. RIWM instead supplies decision-relevant information, such as conflicts, occlusions, interactions, recovery space, and evidence sufficiency, that these mechanisms can translate into constraints, action filters, or fallback conditions.

Probability calibration alone cannot resolve structural epistemic uncertainty arising from missing variables, omitted relations, or incorrect model structure. A hazard outside the represented hypothesis space cannot be recovered by better calibration. RIWM must therefore expose model boundaries and monitor whether critical assumptions remain valid at runtime. Safety conclusions should consequently remain conditional on assumptions about dynamics, state estimation, uncertainty bounds, and operating conditions. RIWM does not replace assurance mechanisms; it evaluates whether their premises are supported by current evidence and triggers revision, active sensing, deferral, or abstention when they are not. World-model output becomes safety-relevant only when its applicability and recovery conditions can be checked and translated into enforceable constraints.

\begin{revisiontext}\label{rev:evidence}
Operationally, sufficiency depends on the proposed action, horizon, and operating domain. Checks assess observation freshness, model applicability, supported risk bounds, and backup feasibility. The world model passes consequences, uncertainty, and unresolved premises to a safety filter, which accepts, modifies, or rejects the proposed action under its own assumptions. Execution feedback updates these checks and indicates whether further sensing or revised control is needed.
\end{revisiontext}

\subsection{\added{Illustrative applications}}

\begin{revisiontext}\label{rev:workflow}
\textbf{Autonomous driving.} At an unprotected left turn, an autonomous vehicle could use six steps: (1) represent occlusion and stopping margin; (2) retrieve near misses with comparable interactions; (3) compare waiting, creeping, and turning; (4) assess conflicts, recovery options, and supporting evidence; (5) execute a supported turn (act), adjust timing (revise), creep safely for visibility (sense), wait where feasible (defer), or abandon the turn using a feasible backup (abstain); (6) reassess as observations arrive and update memory from outcomes.
\end{revisiontext}

\begin{revisiontext}\label{rev:robot}
\textbf{Humanoid locomotion.} World Model Reconstruction \citep{sun2025wmr} estimates contact, foot friction, and payload. For a loaded humanoid on uncertain-friction ground, RIWM could use the same six steps: (1) represent contact and balance; (2) retrieve comparable slips; (3) compare steps and routes; (4) assess balance loss, stabilizing control, and supporting evidence; (5) step (act), change route (revise), probe contact (sense), wait after stabilizing (defer), or abandon the step while maintaining balance (abstain); (6) reassess as observations arrive and update memory from motion outcomes. Because load changes balance and recovery conditions, waiting may require adjusting support first.
\end{revisiontext}

\section{Open challenges and outlook}

Translating RIWM from a conceptual framework into a practical paradigm requires progress in how consequential evidence is defined, tested, retained, and connected to executable safety mechanisms.

\subsection{Open challenges}

\textbf{What constitutes a decision-relevant consequence?} Safety-critical decisions involve physical harm, social interaction, comfort, rule compliance, task completion, and recoverability. These dimensions cannot always be reduced to a single scalar risk score. A key challenge is to determine which consequential differences are sufficient to change an action and how heterogeneous forms of evidence and constraints should be combined.

\needspace{3\baselineskip}
\begin{revisiontext}\label{rev:evaluation}
\textbf{How should decision relevance be evaluated?} Evaluation should test sensitivity to action-changing evidence and robustness to nuisance changes. Candidate measures include critical-branch recall, decision-change recall, action-ranking robustness, recovery success, and unsafe acceptance versus unnecessary deferral. Comparisons should match information and computation budgets. World-model-based scenario generation \citep{zhang2026longtail} motivates targeted variations in agent behavior and environmental conditions, allowing these measures to be examined across successive decisions.
\end{revisiontext}

\textbf{When does action conditioning support counterfactual reasoning?} Action-conditioned generation is not equivalent to reliable intervention reasoning. Observational data record outcomes only for executed actions, while responses to unchosen actions remain unobserved. Counterfactual claims therefore require explicit causal assumptions, uncertainty bounds, and distributional limits, supported where possible by interactive simulation, controlled experiments, natural experiments, or learned response models.

\textbf{When should episodic memory be trusted?} Past experience may become invalid under changes in sensing, policy, traffic rules, or operating conditions, whereas excessive forgetting may cause repeated failures. Memory provenance, confidence decay, conflict tracking, versioning, and revalidation are therefore central challenges for long-term safety memory.

\textbf{How can learned consequences become enforceable constraints?} RIWM may represent consequences in semantic or generative form, whereas runtime assurance typically operates on states, reachable sets, constraints, or verifiable temporal properties. Establishing a reliable interface between these representations while preserving uncertainty and applicability conditions remains a \mbox{major challenge.}

\textbf{When is there enough evidence to act?} Safety-critical systems must avoid both underestimating long-tail risk and becoming paralyzed by unbounded hazard enumeration. The key question is whether available evidence supports a recoverable decision. Computation and caution should therefore adapt to consequence, epistemic uncertainty, recovery margin, and information value, while allowing the system to act, revise, sense, defer, or abstain.

\subsection{Outlook}

Future world models will likely achieve longer horizons, stronger controllability, and greater visual consistency. Their value in safety-critical embodied systems, however, will depend less on generating ever more futures than on recognizing the consequential ones, especially when candidate actions induce different responses, evidence is incomplete, or recovery space is narrowing.

This perspective calls for a shift from prediction-centric to decision-centric world modeling. Risk should not be appended after prediction; it should shape what the model represents, reasons about, and remembers. Decision-Relevant Representation should preserve information that changes action boundaries; Counterfactual Reasoning should focus finite computation on credible alternatives; Safety-Critical Episodic Memory should allow relevant experience to revise current judgment; and Runtime Safety Assurance should connect these judgments to \mbox{executable constraints.}

In the longer term, safety-critical world modeling may progress from predicting what may happen, to identifying what could change the action, and ultimately to recognizing when the available evidence is insufficient to act. The next frontier may therefore lie not in imagining more futures, but in identifying which futures matter, revising judgment through relevant experience, and recognizing when to act, sense, defer, or abstain.


\section*{Acknowledgements}
This research was conducted by KAIST as part of a joint research project under the Korea Institute of Science and Technology Information (KISTI) R\&D program, ``Development of the Next-Generation Integrated Wired/Wireless Communication Gateway (X-Gateway).''


\bibliographystyle{riwm-jicv}
\bibliography{references}
\end{multicols}
\clearpage
\vspace{10pt}
\setlength{\multicolsep}{0pt}
\setlength{\smallskipamount}{2pt}
\begin{multicols}{2}

\begin{biography}{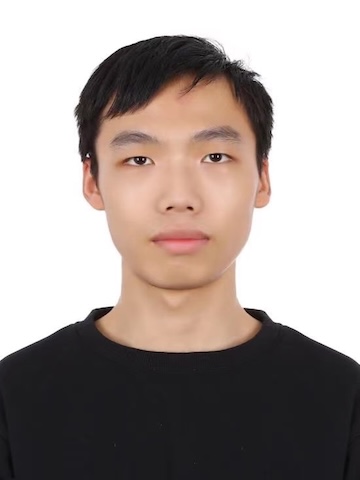}
\textbf{Kailang Ma} received the B.S. degree in information security and the M.S. degree in cyberspace security from Beihang University, Beijing, China, in 2021 and 2024, respectively. He was an Algorithm Engineer at WeRide, an autonomous driving company, from 2024 to 2026. He is currently pursuing the Ph.D. degree with the Cho Chun Shik Graduate School of Mobility, Korea Advanced Institute of Science and Technology (KAIST), under the supervision of Prof. Heye Huang. His research interests include safe and trustworthy autonomy, generative AI for autonomous driving, world models, and risk-sensitive decision making.
\end{biography}

\begin{biography}{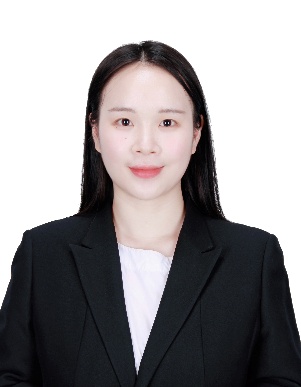}
\textbf{Heye Huang} received the Ph.D. degree in mechanical engineering from Tsinghua University in 2023. She was a Postdoctoral Associate at MIT SMART center and UW--Madison, in 2025 and 2023, respectively. She is currently an Assistant Professor with the Cho Chun Shik Graduate School of Mobility, Korea Advanced Institute of Science and Technology. Her current research interests include safe and trustworthy autonomy, generative AI, risk-sensitive decision making, and human-centered AI for autonomous systems.
\end{biography}

\begin{biography}{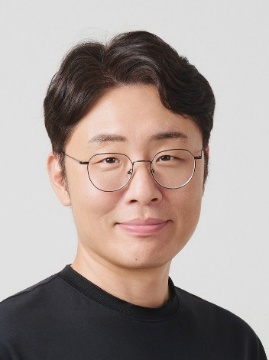}
\textbf{Inhi Kim} received the Ph.D. degree from the Department of Civil Engineering, The University of Queensland, in 2014. He was a Lecturer and later a Senior Lecturer with the Department of Civil Engineering, Monash University, Australia. He is currently an associate professor with the Cho Chun Shik Graduate School of Mobility, KAIST. His current research interests include traffic planning and operation, digital twins, traffic simulation, and the safety and efficiency evaluation of connected vehicles.
\end{biography}

\begin{biography}{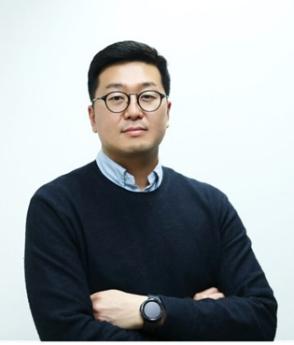}
\textbf{Kitae Jang} received the Ph.D. degree in civil and environmental engineering from the University of California, Berkeley, in 2011. He is currently a Professor with the Cho Chun Shik Graduate School of Mobility, KAIST. He also serves as the Dean of KAIST Academy and the Director of the Center for Excellence in Learning and Teaching and the Research Center for Eco-friendly and Smart Vehicles. His current research interests include transportation energy and environmental policy, traffic operations, intelligent transportation systems, and transportation systems for healthy and safe communities.
\end{biography}